\documentclass[letterpaper, 10 pt, conference]{ieeeconf}  

\IEEEoverridecommandlockouts                              

\usepackage{amsmath}
\usepackage{amssymb}
\usepackage{bm}
\usepackage{dsfont}
\usepackage{algorithm, algorithmic}
\usepackage{graphicx}
\usepackage[bookmarks=true]{hyperref}
\usepackage{subcaption}
\usepackage{cleveref}

\newtheorem{theorem}{Theorem}[section]

\title{\LARGE \bf
Efficient Exploration Using Extra Safety Budget in Constrained Policy Optimization
}

\author{Haotian Xu$^{1,*}$, Shengjie Wang$^{2,*}$, Zhaolei Wang$^{3}$, Yunzhe Zhang$^{2}$, Qing Zhuo$^{1}$,  \\ Yang Gao$^{2,\dag}$ and Tao Zhang$^{1,\dag}$, \textit {Senior Member, IEEE}
\thanks{* Equal contribution. $\dag$ Corresponding author: taozhang@tsinghua.edu.cn, gaoyangiiis@mail.tsinghua.edu.cn.}
\thanks{$^{1}$Department of Automation, Tsinghua University. $^{2}$Department of IIIS, Tsinghua University. $^{3}$Beijing Aerospace Automatic Control Institute.}%
\thanks{This research was funded by Scientific and Technological Innovation 2030 under Grant 2021ZD0110900 and partially supported by the National Natural Science Foundation of China under Grant U21B6002.}
}

\begin{document}

\maketitle
\thispagestyle{empty}
\pagestyle{empty}

\begin{abstract}

Reinforcement learning (RL) has achieved promising results on most robotic control tasks. Safety of learning-based controllers is an essential notion of ensuring the effectiveness of the controllers. Current methods adopt whole consistency constraints during the training, thus resulting in inefficient exploration in the early stage. 
In this paper, we propose an algorithm named Constrained Policy Optimization with Extra Safety Budget (ESB-CPO) to strike a balance between the exploration efficiency and the constraints satisfaction. In the early stage, our method loosens the practical constraints of unsafe transitions (adding extra safety budget) with the aid of a new metric we propose. With the training process, the constraints in our optimization problem become tighter. Meanwhile, theoretical analysis and practical experiments demonstrate that our method gradually meets the cost limit's demand in the final training stage. When evaluated on Safety-Gym and Bullet-Safety-Gym benchmarks, our method has shown its advantages over baseline algorithms in terms of safety and optimality. Remarkably, our method gains remarkable performance improvement under the same cost limit compared with baselines.

\end{abstract}

\section{INTRODUCTION}
Reinforcement learning (RL) has shown great promise in many robotic control tasks \cite{hwangbo2019learning,andrychowicz2020learning}. RL-based algorithm can facilitate the agent to maximize the expected sum of rewards (return), which is a manually designed metric. Meanwhile, safety of the agent should be considered carefully due to the existing obstacles or other constraints in the real-world applications \cite{garcia2015comprehensive,yang2022safe}. Therefore, reaching the optimality and safety remains an essential problem in the field of RL.



\begin{figure}[t!]
    \centering
    \includegraphics[width=0.9\linewidth]{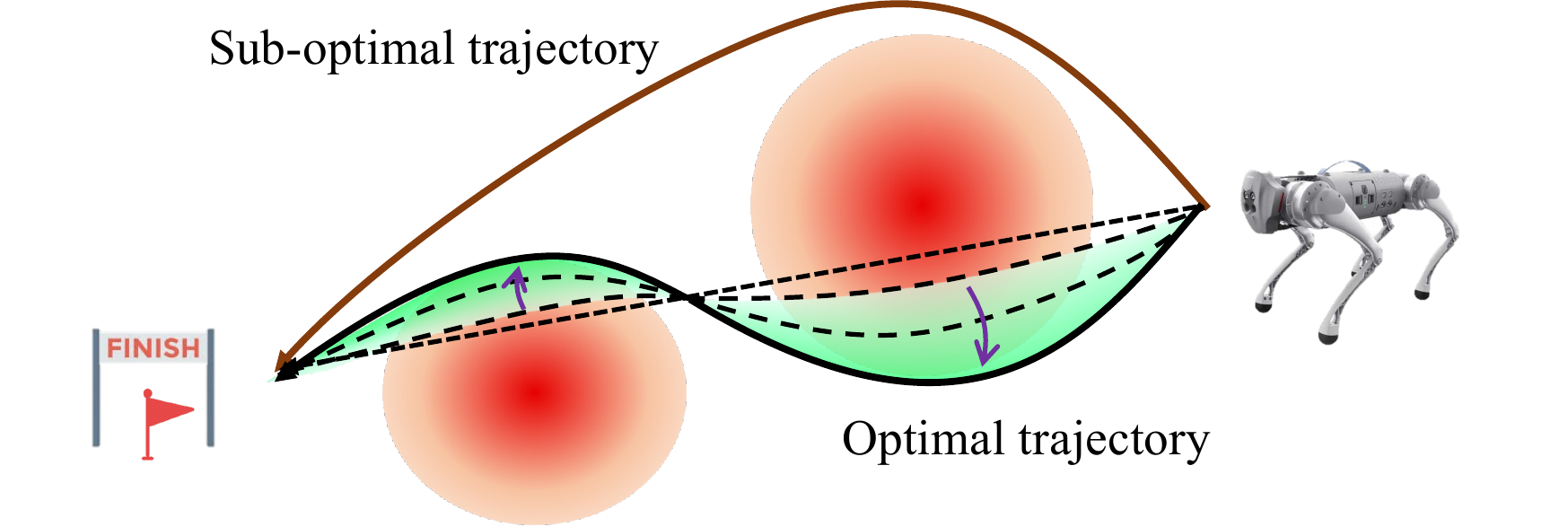}
    \caption{Intuitive example showing the impact of efficient exploration in the early stage. The red regions represent the obstacles. When the robot concerns the safety constraints a lot at the initial stage, it may find a sub-optimal trajectory. Instead, if the robot ignores the constraints for unsafe states first, it can find a direct path to finish the task. Afterward, it can meet the demand of avoiding collisions gradually so that the optimal trajectory can be finally obtained.}
    \label{fig:toy}
\end{figure}

To balance rewards and costs, researchers propose Lagrangian-based methods to transfer the prime problem to an unconstrained problem with Lagrangian multipliers \cite{stooke2020responsive,liu2020ipo}. Because those methods lack theoretical guarantee, safe RL algorithms based on trust region method are proposed to achieve the adequate policy updating \cite{achiam2017constrained,Yang2020Projection-Based}. Furthermore, inspired by the success of Lyapunov stability theorem, there exists some methods that design safety certificates to ensure constraint satisfaction \cite{chow2018lyapunov,yang2023model}. Despite the recent progress, previous studies adopt the whole consistency constraints in the training process. The strong constraints are likely to harm the policy exploration in the early training stage. Thus, whole consistency constraints are detrimental to the early exploration and the policy may be trapped into the sub-optimal points. 

To address this issue, the proposed algorithm should encourage early exploration and gradually maintain safety constraints. We provide a toy example shown in Fig. \ref{fig:toy}, to illustrate why the above strategy can improve the performance. Based on the simple yet effective idea, we propose Constrained Policy Optimization with Extra Safety Budget (ESB-CPO) algorithm \footnote{See our project page at \href{https://sites.google.com/view/esb-cpo}{https://sites.google.com/view/esb-cpo}.}. Our method can achieve higher rewards under the same cost limits compared with baselines.   
Our contribution can be summarized as follows:
\begin{itemize}
\item We construct a novel metric, Lyapunov-based Advantage Estimation (LAE), to evaluate the safe and unsafe transitions. It consists of two parts, stability value and safety value. Safety value part has a significant impact only on unsafe transitions. 
\item We propose Constrained Policy Optimization with Extra Safety Budget (ESB-CPO) algorithm based on LAE. To encourage exploration, our method loosens the constraints of unsafe transitions by adding an extra safety budget which comes from the safety value part of LAE. Furthermore, the extra safety budget becomes very close to 0 in the final stage of training.
\item To achieve the goal in ESB-CPO, we update the two factors, $\alpha$ and $\beta$, in LAE using the adaptive methods. By introducing a variable concerning safety, LAE can distinguish safe and unsafe transitions via $\beta$. In the early stages, the optimization-based adaptation of $\alpha$ controls the degree to which constraints are loosened.
\end{itemize}

\section{RELATED WORKS}
Safe reinforcement learning aims to solve a constrained optimization problem with safety constraints. Constrained Markov Decision Process (CMDP) is commonly used to describe this problem. Concretely, the safe policy can satisfy the expected sum of safety violation costs below a given threshold. Some methods transform such constrained problem into an unconstrained problem \cite{liu2021policy,ding2021provably}. Previous methods introduced Lagrangian relaxation to take the rewards and costs into consideration together \cite{stooke2020responsive,Ray2019,peng2022model}. Liu \emph{et al.} applied the original dual interior point method to constrained reinforcement learning, transformed the constraints into the penalty of the objective function by using logarithmic barrier function \cite{liu2020ipo}. 
Another line of work added trust region constraints in policy optimization, thus providing a guarantee of safety violations \cite{achiam2017constrained}. TRPO is a model-free RL-based algorithm to guarantee the monotonicity of policy updating \cite{schulman2015trust}. CPO was proposed to implement constrained reinforcement learning based on TRPO \cite{achiam2017constrained}. Chow \emph{et al.} proposed an algorithm based on Lyapunov function to ensure safety during the training process \cite{chow2018lyapunov,chow2019lyapunovbased}. Inspired by traditional control methods, some researchers proposed algorithms to jointly learn a policy and a neural barrier certificate under stepwise state constraint setting \cite{yang2023model,mathiesen2022safety}. Furthermore, some researchers used a two-stage method to ensure safety at each time step \cite{Yang2020Projection-Based,yang2022cup,xu2020primal,srinivasan2020learning}. The first stage of each training step uses TRPO to solve the unconstrained optimization problem, and the second stage projects the policy from the first stage onto the policy that satisfies the constraints. Additionally, introducing new variables that directly reflect the current state's safety is also a promising direction \cite{sootla2022saute, sootla2022enhancing}. A. Sootla \emph{et al.} proposed Sauté RL, which uses a new state that reflects the current safety of the system as the cumulative loss changes and reflects it in the reward function, so that the agent can well satisfy the constraints \cite{sootla2022saute}. However, those methods use the whole consistency constraints, thus making the agent learn a comparatively conservative policy. Our method encourages the agent to explore in the early stage, then restrict the agent's behaviour gradually until the agent satisfies the safety constraints. 


\section{PRELIMINARY}\label{pre}
MDP (Markov Decision Process) is defined as a tuple ($\mathcal S$, $\mathcal A$, $\mathcal P$, $\mathcal R$, $\mathcal \mu$, $\mathcal \gamma$), where $\mathcal S$ and $\mathcal A$ are the state and action space respectively, $\mathcal P: \mathcal S \times \mathcal A \times \mathcal S \mapsto [0, 1]$ is the transition probability function, $\mathcal R: \mathcal S \times \mathcal A \times \mathcal S \mapsto \mathds{R}$ is the reward function, $\mathcal \mu: \mathcal S \mapsto [0, 1]$ is the distribution of initial state, $\mathcal \gamma$ is the discount factor for future rewards. CMDP is defined as a tuple ($\mathcal S$, $\mathcal A$, $\mathcal P$, $\mathcal R$, $\mathcal C_i$, $\mathcal \mu$, $\mathcal \gamma$), where $\mathcal S$, $\mathcal A$, $\mathcal P$, $\mathcal R$ and $\mathcal \mu$ have the same meanings as in MDP, $\mathcal C_i: \mathcal S \times \mathcal A \times \mathcal S \mapsto [0, +\infty)$ is the cost function which describe the satisfaction of the $i$-th constraint, $\mathcal \gamma$ is the discount factor for both future rewards and costs.

A policy $\mathcal \pi: \mathcal S \mapsto P(\mathcal A)$ maps given states to probability distributions over action space and $\pi(a_t|s_t)$ is the probability of taking action $a$ under state $s$ in time step $t$. We use $\pi_\theta$ to describe a policy parameterized by $\theta$. 
The expected discounted cumulative return of a policy is
\begin{equation}
J^R(\theta)=\mathds{E}_{\tau \sim \pi_\theta}\left[ \sum_{t=0}^{\infty}\gamma^t R(s_t, a_t, s_{t+1})\right]\label{JR}
\end{equation} where $\tau \sim \pi_\theta$ is a trajectory sampled from $\pi_\theta$. The expected discounted cumulative cost of a policy is
\begin{equation}
J^{C_i}(\theta)=\mathds{E}_{\tau \sim \pi_\theta}\left[ \sum_{t=0}^{\infty}\gamma^t {C_i}(s_t, a_t, s_{t + 1})\right]\label{JC}
\end{equation}
The optimization goal of a safe RL algorithm is to find the optimal policy $\pi_{\theta^*}$ which maximizes $J^R$ while guarantees $J^{C_i} \leq d_i$, where $d_i$ is the cost limit for the $i$-th constraint. Formally, the optimization problem is defined as:
\begin{equation}
\begin{aligned}
    & \max_\theta J^R(\theta) \\
    & {\rm s.t.} \quad J^{C_i}(\theta) \leq d_i
\end{aligned}
\label{problem defination}
\end{equation}

Standard definitions of the value function $V_{\theta}$, the state-action value function $Q_{\theta}$, the cost value function $V_{\theta}^{C_i}$ and the state-action cost value function $Q_{\theta}^{C_i}$ are exploited in most previous studies. Thus we omit them. The commonly used advantage functions are defined as $A_{\theta}^R(s,a)=Q_{\theta}(s,a)-V_{\theta}(s)$ and $A_{\theta}^{C_i}(s,a)=Q_{\theta}^{C_i}(s,a)-V_{\theta}^{C_i}(s)$.



The following theorem provides bounds of the error of the objectives and constraints with $\pi_\theta$ and $\pi_{\Tilde{\theta}}$ \cite{achiam2017constrained}.  

\begin{theorem}
    For any function $f: \mathcal S \times \mathcal A \times \mathcal S \mapsto \mathds{R} $ and any policies $\pi_\theta$ and $\pi_{\Tilde{\theta}}$, define
    \begin{equation}
        \epsilon_f^{\Tilde{\theta}} \doteq \max_{s_t} \left| \mathop{\mathds{E}}_{a_t \sim \pi_{\Tilde{\theta}} \atop s_{t+1} \sim P_{a_t}^{s_t}} \left[ f(s_t, a_t, s_{t+1}) \right] \right|, \label{epsilon}
    \end{equation}
    The following bounds hold:
    \begin{equation}
        \begin{aligned}
            J^R(\Tilde{\theta})
                \geq &J^R(\theta) + \frac{1}{(1-\gamma)}\mathop{\mathds{E}}_{s \sim \rho_\theta \atop a \sim \pi_{\theta}} \left[  \frac {\pi_{\Tilde{\theta}}(a|s)}{\pi_{\theta}(a|s)} A_\theta^R (s,a)\right] \\
                &- M_\theta^R (\Tilde{\theta}) \mathop{\mathds{E}}_{s \sim \rho_\theta} \left[ D_{TV}(\Tilde{\theta} || \theta) [s] \right],
            \label{Jrbound}
        \end{aligned}
    \end{equation}
    \begin{equation}
        \begin{aligned}
            J^{C_i}(\Tilde{\theta}) 
                \leq &J^{C_i}(\theta) + \frac{1}{(1-\gamma)}\mathop{\mathds{E}}_{s \sim \rho_\theta \atop a \sim \pi_{\theta}} \left[  \frac {\pi_{\Tilde{\theta}}(a|s)}{\pi_{\theta}(a|s)} A_\theta^{C_i} (s,a)\right] \\
                &+ M_\theta^{C_i} (\Tilde{\theta}) \mathop{\mathds{E}}_{s \sim \rho_\theta} \left[ D_{TV}(\Tilde{\theta} || \theta) [s] \right],
            \label{Jcbound}
        \end{aligned}
    \end{equation}
    where $M^{{R}}_{\theta}(\Tilde{\theta})=\frac{2 \gamma }{(1-\gamma)^2} \epsilon_{V_\theta}^{\Tilde{\theta}}$, $M^{{C_i}}_{\theta}(\Tilde{\theta})=\frac{2 \gamma }{(1-\gamma)^2} \epsilon_{V_\theta^{C_i}}^{\Tilde{\theta}}$.
\end{theorem}

These bounds can be used as surrogate objectives to guarantee theoretically monotonic improvement in policy search update. 
CPO \cite{achiam2017constrained} is a practical algorithm using these surrogate objectives with trust region theorems, which optimizes (\ref{problem defination}) by following update step: 
\begin{equation}
\begin{aligned}
    \theta' = & \mathop{\rm argmax}_{\Tilde{\theta}}  \mathop{\mathds{E}}_{s \sim \rho_\theta \atop a \sim \pi_{\theta}}\left[ \frac {\pi_{\Tilde{\theta}}(a|s)}{\pi_{\theta}(a|s)} A_{\theta}^R (s, a) \right] \\
    {\rm s.t.} \quad 
    & J^{C_i}(\theta) + \frac{1}{(1-\gamma)}\mathop{\mathds{E}}_{s \sim \rho_\theta \atop a \sim \pi_{\theta}}\left[ \frac {\pi_{\Tilde{\theta}}(a|s)}{\pi_{\theta}(a|s)} {A_{\theta}^{C_i}} (s) \right] \leq d_i \\
    & \mathop{\mathds{E}}_{s \sim \rho_\theta} \left[ D_{KL}(\pi_{\Tilde{\theta}}(\cdot | s) || \pi_{\theta}(\cdot | s)) \right] \leq \delta 
    \label{CPO}
\end{aligned}
\end{equation}

\section{METHODOLOGY} 

\subsection{Lyapunov-Based Advantage Estimation}

The existence of Lyapunov function becomes an effective tool to evaluate the system's stability in RL \cite{chang2021stabilizing,han2020actor,wang2023rl}. Recent studies utilized the control Lyapunov function (CLF) to assess the system's safety, achieving promising results on some robotic tasks \cite{lawrence2020almost,mittal2020neural,dawson2022safe}. 
Inspired by those successes, we find Lyapunov function and CLF can separately evaluate the performance of safe and unsafe transitions. Thus, we construct a new metric, namely Lyapunov-based Advantage Estimation (LAE) ${A^{C_i}_\theta}'(s_t, a_t)$ as follows. 
\begin{equation}
\begin{aligned}
    {A^{C_i}_\theta}'(s, a) &=  
    \mathop{\mathds{E}}_{s' \sim P_a^{s}} [ V^{C_i}_\theta(s') -V^{C_i}_\theta(s)  \\  &+ \alpha \left( V^{C_i}_\theta(s) -  \beta V^{C_i}_\theta(s') \right) ] 
    \label{Lya adv}
\end{aligned}
\end{equation}
where $\alpha \in (0, 1), \beta \in [0, 1]$ are adaptive factors, $s_{t}$, $a_{t}$ and $s_{t+1}$ are marked as $s$, $a$ and $s'$ respectively. Furthermore, $P_a^{s}$ is the distribution of the next state after $s_t$ with $a_t$.

We can notice that when ${A^{C_i}_\theta}'(s,a) \leq 0$, that means 
\begin{equation}
\begin{aligned}
    &\mathop{\mathds{E}}_{s' \sim P_a^{s}} \left[ V^{C_i}_\theta(s') \right] -V^{C_i}_\theta(s) \\
    & \leq -\alpha \left( V^{C_i}_\theta(s) -  \beta \mathop{\mathds{E}}_{s' \sim P_a^{s}} \left[ V^{C_i}_\theta(s') \right] \right)
    \label{constraint function}
\end{aligned}
\end{equation}

\begin{figure}[t!]
    \centering
    \includegraphics[width=0.25\textwidth]{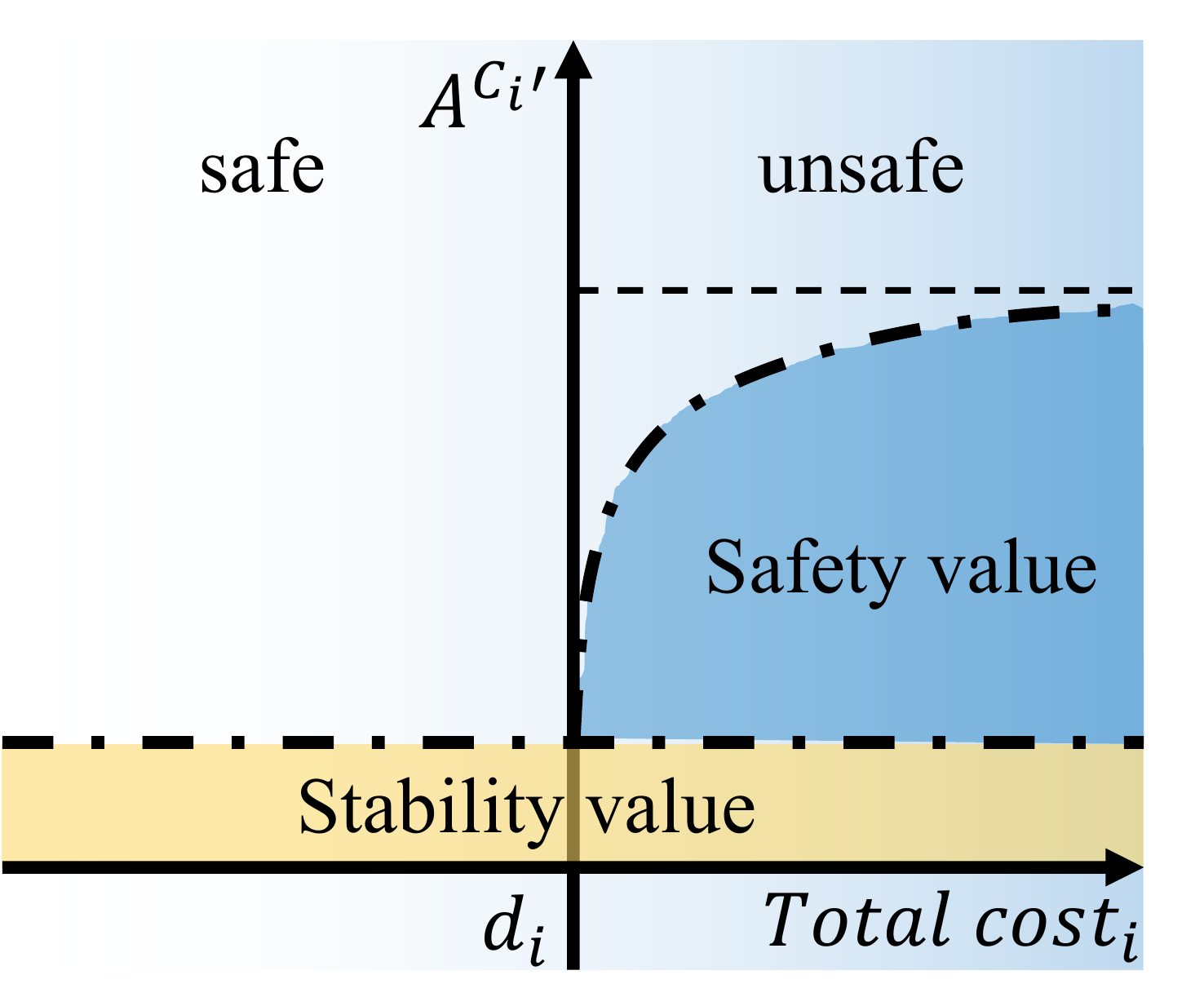}
    \caption{Stability value and safety value of Lyapunov-based advantage (LAE) under safe and unsafe transitions.}
    \label{fig:adv}
\end{figure}

Concretely, the above inequality corresponds to the Lyapunov function constraints when  $\beta = 1$ holds \cite{murray2017mathematical}. On the other hand, when $\beta = 0$ holds, it equals to the constraints of a control Lyapunov function (CLF) \cite{mittal2020neural}. Intuitively, CLF is a stronger constraint than the Lyapunov function due to containing an extra safety consideration. Fig. \ref{fig:adv} shows an illustrative example to depict the relationship between our advantage estimation ${A^{C_i}_\theta}'(s, a)$ and the total cost. 
${A^{C_i}_\theta}'(s, a)$ contains two parts, stability and safety values.  
In the safe region, it only represents a value function concerning stability. When the agent is in unsafe region, extra part is similar to a metric of safety. This indicates that LAE evaluates safe transitions' performance depends on the stability value part, while evaluates unsafe transitions depends on the stability and safety value parts. $\beta$ adjusts the evaluation. It controls the threatening estimation of the transition according to the policy's satisfaction of safety at step $t$. To sum up, our advantage estimation can magnify the gap between safe and unsafe transitions by safety value.

\subsection{Constrained Policy Optimization With Extra Safety Budget}
Based on the problem \eqref{CPO}, we derive our optimization problem using LAE, which updates policy as:
\begin{equation}
\begin{aligned}
    \theta' = & \mathop{\rm argmax}_{\Tilde{\theta}}  \mathop{\mathds{E}}_{s \sim \rho_\theta \atop a \sim \pi_{\theta}}\left[ \frac {\pi_{\Tilde{\theta}}(a|s)}{\pi_{\theta}(a|s)} A_{\theta}^R (s, a) \right] \\
    {\rm s.t.} \quad 
    & J^{C_i}(\theta) + \frac{1}{(1-\gamma)}\mathop{\mathds{E}}_{s \sim \rho_\theta \atop a \sim \pi_{\theta}}\left[ \Delta_{\theta, \Tilde{\theta}} (s,a) \frac{{A_{\theta}^{C_i}}' (s, a)}{1 - \alpha_{i\theta}} \right] \leq d_i \\
    & \mathop{\mathds{E}}_{s \sim \rho_\theta} \left[ D_{KL}(\pi_{\Tilde{\theta}}(\cdot | s) || \pi_{\theta}(\cdot | s)) \right] \leq \delta 
    \label{ESB-CPO}
\end{aligned}
\end{equation}
where $\alpha_{i\theta}$ decreases from $1^-$ to $0$ with updating and $\Delta_{\theta, \Tilde{\theta}} (s, a) = \frac {\pi_{\Tilde{\theta}}(a|s)}{\pi_{\theta}(a|s)} - 1$. $\Delta_{\theta, \Tilde{\theta}} (s, a)$ describes the tendency of the policy to be updated from $\pi_\theta$ to $\pi_{\Tilde{\theta}}$. If the new policy try to avoid choosing action $a$ under $s$, $\Delta_{\theta, \Tilde{\theta}} (s, a) < 0$; on the contrary, $\Delta_{\theta, \Tilde{\theta}} (s, a) > 0$. In the following part, we will introduce why our method can encourage exploration in the early stage and meet the demand of safety gradually. 

First, we can get the relationship between ${A_{\theta}^{C_i}}' (s, a)$ and $A_{\theta}^{C_i}(s,a)$:
\begin{equation}
    \frac{{A_{\theta}^{C_i}}' (s, a)}{1-\alpha_{i\theta}} = A_{\theta}^{C_i}(s,a) + B_{1\theta}^i (s, a) + B_{2\theta}^i (s'),
\end{equation}
where $B_{1\theta}^i (s, a) = (1-\gamma)V_\theta^{C_i}(s') - {C_i}(s, a, s')$, $B_{2\theta}^i (s) = \frac{\alpha_{i\theta} (1 - \beta_{i\theta} (s))}{1-\alpha_{i\theta}} V_\theta^{C_i}(s)$. Therefore, (\ref{ESB-CPO}) can be obtained by adding two gaps in the constraint function in (\ref{CPO}):
\begin{equation}
    \begin{aligned}
        &J^{C_i}(\theta) + \frac{1}{(1-\gamma)}\mathop{\mathds{E}}_{s \sim \rho_\theta \atop a \sim \pi_{\theta}}\left[ \Delta_{\theta, \Tilde{\theta}} (s,a) \frac{{A_{\theta}^{C_i}}' (s, a)}{1 - \alpha_{i\theta}} \right] \leq d_i \\
       &\Leftrightarrow 
       \begin{aligned}
           &J^{C_i}(\theta) + \frac{1}{(1-\gamma)}\mathop{\mathds{E}}_{s \sim \rho_\theta \atop a \sim \pi_{\theta}}\left[ \frac {\pi_{\Tilde{\theta}}(a|s)}{\pi_{\theta}(a|s)} {A_{\theta}^{C_i}} (s) \right] \\
            &+ G_{1\theta}^i (s, a) + G_{2\theta}^i (s, a)\leq d_i
       \end{aligned}
    \end{aligned}
\end{equation}
where $G_{1\theta}^i (s, a) = \frac{1}{1-\gamma}\mathop{\mathds{E}}_{s \sim \rho_\theta \atop a \sim \pi_{\theta}}\left[ \Delta_{\theta, \Tilde{\theta}} (s,a) B_{1\theta}^i (s, a) \right]$, $G_{2\theta}^i (s, a) = \frac{1}{1-\gamma}\mathop{\mathds{E}}_{s \sim \rho_\theta \atop a \sim \pi_{\theta}}\left[ \Delta_{\theta, \Tilde{\theta}} (s,a) B_{2\theta}^i (s') \right]$.

It's clear that if these gaps are negative, they loosen the constraint, otherwise they tighten the constraint. Therefore, considering the safety budget is defined as $d_i - J_{C_i}(\theta)$, the gaps can be seen as Extra Safety Budgets (ESBs). Notice that if gap is negative, the corresponding ESB is positive:
\begin{equation}
    \rm ESB = -Gap.
\end{equation}
 In the rest of this subsection, we'll show how these ESBs match our new metric and how they influence the policy update in detail.

\subsubsection{ESB For Stability}
Notice that 
\begin{equation}
    A_{\theta}^{C_i}(s,a) + B_{1\theta}^i (s, a) = V_{\theta}^{C_i}(s') - V_{\theta}^{C_i}(s). 
\end{equation}

This means by adding $G_{1\theta}^i (s, a)$ we actually use our stability value replacing the advantage function. Our experiments demonstrate that $G_{1\theta}^i (s, a)$ is very close to 0 in practical usage. 
Furthermore, $B_{1\theta}^i (s, a)$ provides another advantage. When a transition is safe, a sparse-costs environment will give zero immediate cost. Thus $B_{1\theta}^i (s, a)$ can provide a prediction of average future cost. 
Since $\gamma < 1$, $B_{1\theta}^i (s, a)$ is always positive. 
It means we tighten the bound when $\Delta_{\theta, \Tilde{\theta}} (s,a) > 0$ (the action should be encouraged). 


\subsubsection{ESB Balancing Exploration Efficiency and Constraint Satisfaction}
$G_{2\theta}^i (s, a)$ matches the safety value in LAE. From the form of $B_{2\theta}^i (s)$ we can know that, when state $s$ is safe, $\beta_{i\theta} (s) = 1$, thus $B_{2\theta}^i (s) = 0$; when state $s$ is unsafe, $\beta_{i\theta} (s) < 1$, $B_{2\theta}^i (s) > 0$ since $V_\theta^{C_i}(s)$ is positive. 
Similar to $G_{1\theta}^i (s, a)$, if the new policy tends to avoid an unsafe transition in most of states, $G_{2\theta}^i (s, a) < 0$, otherwise $G_{2\theta}^i (s, a)>0$. This means we strengthen the constraint when the policy try to take more risk in most states, while loosen the constraint when the policy tends to be safer. Since the policy is updated to be safer, $G_{2\theta}^i (s, a)$ is more likely to be negative than positive. Thus $G_{2\theta}^i (s, a)$ is more likely to provide positive extra safety budget to loosen the constraint, so that it encourages exploration. 

Furthermore, this encouragement decreases with updating. 
In practical algorithm, $\alpha_{i\theta}$ decreases from $1^-$ towards $0$, thus the influence of $G_{2\theta}^i (s, a)$ becomes weaker.
At the beginning, when $\alpha_{i\theta} \rightarrow 1$, the total constraint will be greatly loosen, thus we can achieve excellent exploration efficiency; with the influence of $G_{2\theta}^i (s, a)$ becomes weaker, the extra safety budget becomes smaller, thus the satisfaction of original constraint can be gradually obtained. Since $G_{2\theta}^i (s, a)$ will be zero when $\alpha_{i\theta}=0$ or $\pi_\theta$ is a safe policy, our policy can reach a satisfaction of constraint no worse than CPO. 

\begin{figure}[t!]
    \centering
    \includegraphics[width=0.5\textwidth]{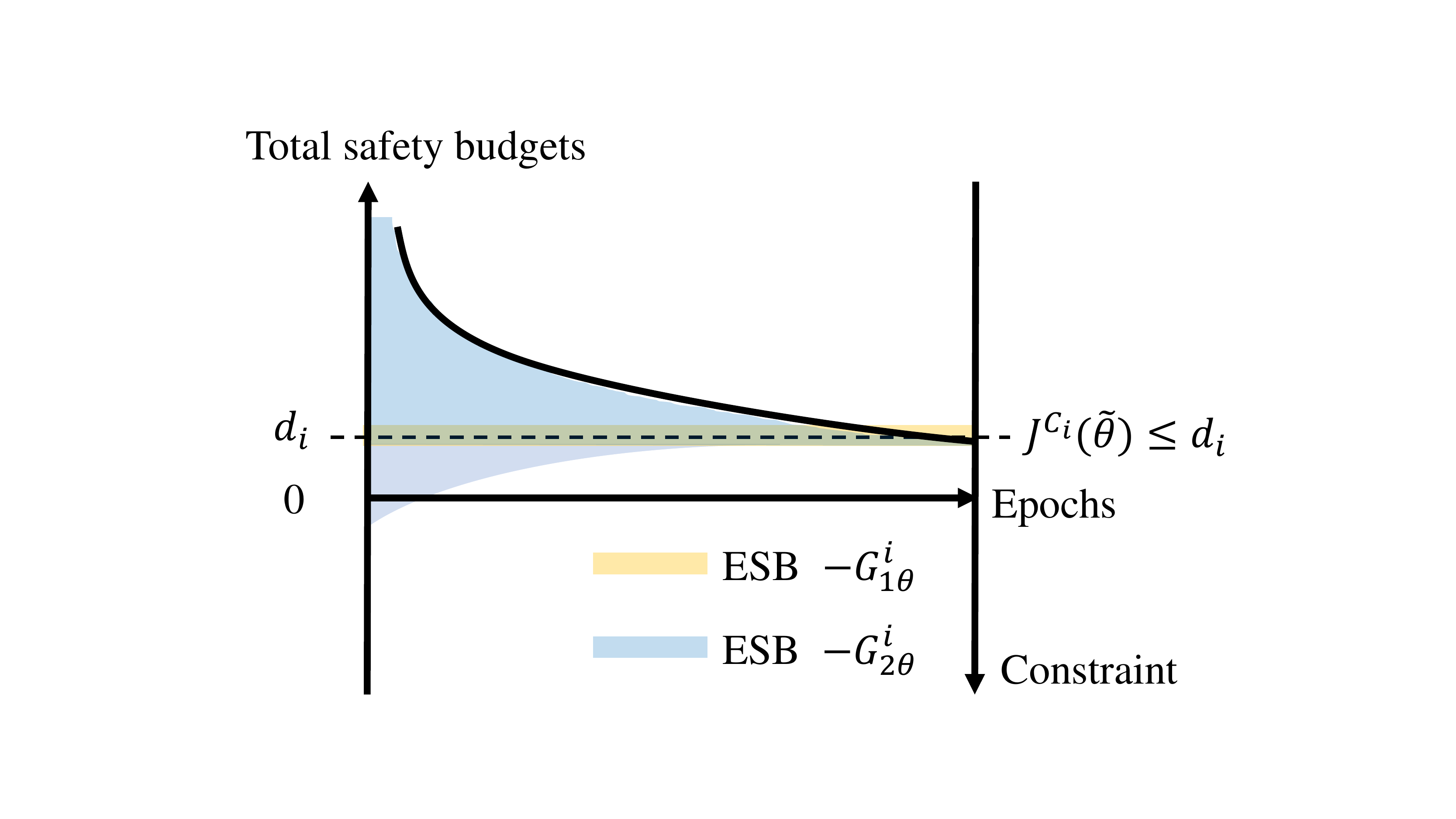}
    \caption{Impact of $G_{1\theta}^i$ and $G_{2\theta}^i$ on practical constraints with time approaching. $G_{1\theta}^i$ is very close to 0 consistently, and the total safety budgets decrease gradually due to the change of $G_{2\theta}^i$. 
    }
    \label{fig:constraint-steps}
\end{figure}

Fig. \ref{fig:constraint-steps} clearly shows the influences of ESBs. The total safety budget determines the constraint: if total safety budget is higher, the constraint is weaker. At the early epochs, $G_{2\theta}^i (s, a)$ provides large extra safety budget w.r.t safety states. In the end, the total safety budget is close to the original cost limit, thus the constraint is close to the original one. $G_{1\theta}^i (s, a)$ provides a small extra safety budget of stability independent to safety or training steps. The direction of policy update determines whether the ESBs are positive or negative.

\subsection{Sample-Based Adaptation of Factors}
According to the above parts, we give practical sample-based methods to update factors $\alpha_{i\theta}$ and $\beta_{i\theta}(s_t)$.
\subsubsection{Adaptation of $\beta_{i\theta}(s_t)$ based on safety states}

A normalized safety state $z_{i\theta}(s_t)$ researchers proposed is a sample-based inner state which directly shows the safety of the state at step $t$, based on the remaining safety budget \cite{sootla2022saute}. The definition of $z_{i\theta}(s_t)$ is
\begin{equation}
    z_{i\theta}(s_t) = \frac{d_i - \sum_{(l=0)}^t \gamma^l {C_i}(s_l, a_l, s_{l+1})}{\gamma^t d_i} ,
\end{equation}
where $s_l$, $a_l$ and $s_{l+1}$ are in trajectory sampled from $\pi_\theta$. Notice that our ESBs are not considered because we need $z_{i\theta}(s_t)$ to show the actual safety. When the sum of costs is lager than the cost limit $d_i$, $z_{i\theta}(s_t)$ is less than 0.

It's easy to find that $z_{i\theta}(s_{t})$ can be updated as
\begin{equation}
    \begin{aligned}
        z_{i\theta}(s_{t+1}) &= \frac{z_{i\theta}(s_t) - \frac{{C_i}(s_t, a_t, s_{t+1})}{d_i}}{\gamma} 
        \label{z_t}
    \end{aligned}
\end{equation}
with an initial value $1$ before $t=0$.

Considering the range $[0,1]$, we calculate $\beta_{i\theta}(s_t)$ by
\begin{equation}
    \beta_{i\theta}(s_t) = 1 + \min \left( \tanh \left( z_{i\theta}(s_t) \right),0 \right). \label{beta}
\end{equation}
When $z_{i\theta}(s_t)$ is less than 0 (unsafe state), $\beta_{i\theta}(s_t)$ decreases towards 0.


\subsubsection{Adaptation of $\alpha_{i\theta}$ based on optimization}

To our concerns, the policy gradient reflects directly how the constraint influencing policy. Therefore, we introduce a Lagrangian multiplier $\lambda_i$ to calculate $\alpha_{i\theta}$ based on the policy gradient of constraint function. 
First, we construct the following local optimization problem:
\begin{equation}
    \min_{\Tilde{\theta}} \max_{\lambda_i} \lambda_i P_{i\theta}(\Tilde{\theta})\label{prime}
\end{equation}
where $P_{i\theta}(\Tilde{\theta})=\mathop{\mathds{E}}_{s \sim \rho_\theta \atop a \sim \pi_{\theta}}\left[ \frac {\pi_{\Tilde{\theta}}(a|s)}{\pi_{\theta}(a|s)} {A_\theta^{C_i}}'(s, a) \right]$. 

The dual problem of (\ref{prime}) is
\begin{equation}
\begin{aligned}
    \max_{\lambda_i} \min_{\Tilde{\theta}} &\quad  \lambda_i P_\theta(\Tilde{\theta}),  \\
    {\rm s.t.} &\quad \lambda_i \geq 0.
\end{aligned}
\end{equation}
Thus $\lambda_i$ can be updated as
\begin{equation}
    \lambda_{i, t+1} = \max \left( \lambda_{i, t} + \eta P_\theta(\Tilde{\theta}), 0 \right),\label{lagragian}
\end{equation}
where $\eta$ is the step size.

Notice that during policy optimization, $P_{i\theta} (\Tilde{\theta})$ is more likely to be negative in the early stage. Therefore, Eq. \ref{lagragian} indicates that $\lambda_i$ decreases with the training process.

Considering the range of $\alpha_{i\theta}$, we calculate $\alpha_{i\theta}$ by
\begin{equation}
    \alpha_{i\theta} = \tanh \left( \frac{k_i}{e^{-\lambda_{i}}} \right) ,
\end{equation}
where $k$ is a hyper parameter which globally controls the decreasing speed of $\alpha_{i\theta}$. As $\lambda_i$ decreases, $\alpha_{i\theta}$ changes from 1 to 0. 



\begin{figure*}[th]
    \centering
    \includegraphics[width=0.88\textwidth]{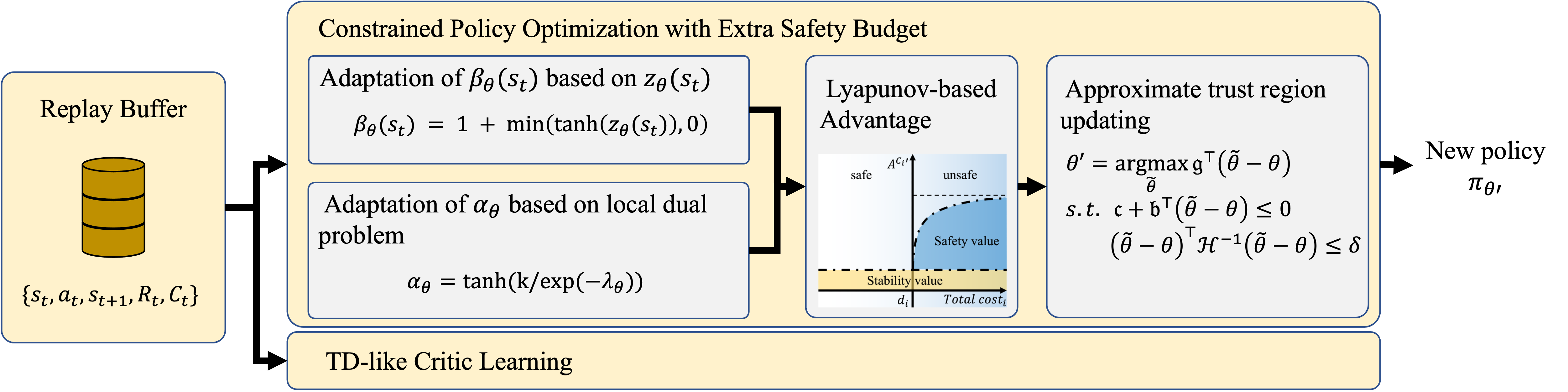}
    \caption{Framework of ESB-CPO algorithm. The method firstly compute the adaptive factors $\alpha_{i\theta}$ and $\beta_{i\theta}(s_t)$. Then LAE value can be obtained by them. Finally we use an approximate trust region method to update the current policy.}
    \label{fig:framework}
\end{figure*}

\subsection{Algorithm Description}

For small step size $\delta$, the optimization problem can be solved approximately by updating with first-order approximation of objective and constraints and second-order approximation of KL-divergence. Denoting the gradient of the objective as $\mathfrak{g}$, the gradient of constraint as $\mathfrak{b}$, the Hessian of the KL-divergence as $\mathcal{H}$, and defining $\mathfrak{c}_i \doteq J^{C_i} (\theta) - d_i$, the approximation to Eq. (\ref{ESB-CPO}) is
\begin{equation}
\begin{aligned}
    \theta' = & \mathop{\rm argmax}_{\Tilde{\theta}}  \mathfrak{g}^\top (\Tilde{\theta} - \theta) \\
    {\rm s.t.} \quad 
    & \mathfrak{c}_i + \mathfrak{b}_i^\top (\Tilde{\theta} - \theta) \leq 0 \\
    & \frac{1}{2}(\Tilde{\theta} - \theta)^\top \mathcal{H} (\Tilde{\theta} - \theta) \leq \delta 
    \label{approx LCPO}
\end{aligned}
\end{equation}

Eq. (\ref{approx LCPO}) directly matches the form of approximate CPO \cite{achiam2017constrained}, whose dual problem is
\begin{equation}
    \max_{\mu_1 \geq 0 \atop \mu_2 \succeq 0} \frac{-1}{2\mu_1}\left( \mathfrak{g}^\top \mathcal{H}^{-1} \mathfrak{g} -2\mathfrak{r}^\top \mu_2 + \mu_2^\top \mathcal{S} \mu_2 \right) + \mu_2^\top c - \frac{\mu_1 \delta}{2},\label{dual approx}
\end{equation}
where $\mathfrak{c} = [\mathfrak{c}_0, \mathfrak{c}_1, ...]$, $\mathfrak{r} \doteq \mathfrak{g}^\top \mathcal{H}^{-1} \mathcal{B}$, $\mathcal{S} \doteq \mathcal{B}^\top \mathcal{H}^{-1} \mathcal{B}$, $\mathcal{B}=[\mathfrak{b}_0, \mathfrak{b}_1, ...]$.

Therefore, in our experiments with a single constraint, Eq. (\ref{approx LCPO}) can be solved via approximate CPO updating:
\begin{equation}
        {\rm If \quad (\ref{dual approx}) \quad is \quad feasible:} \quad \hat{\theta} = \theta + \frac{1}{\mu_1^*}\mathcal{H}^{-1}(\mathfrak{g} - \mu_2^* \mathfrak{b}), \label{feasible}
\end{equation}
\begin{equation}
        {\rm else:} \quad \hat{\theta'} = \theta - \sqrt{\frac{2\delta}{\mathfrak{b}^\top \mathcal{H}^{-1} \mathfrak{b}}} \mathcal{H}^{-1} \mathfrak{b},
        \label{unfeasible}
\end{equation}
where $\mu_1^*$ and $\mu_2^*$ are solutions to (\ref{dual approx}). Finally the new policy $\pi_{\theta'}$ is obtained by backtracking line searching to enforce satisfaction of constraints. 


 The pseudo-code of our algorithm is shown as Algorithm \ref{alg:1}, the corresponding frameworks is shown in Fig. \ref{fig:framework}.

\begin{algorithm}[h]
	\renewcommand{\algorithmicrequire}{\textbf{Input:}}
	\renewcommand{\algorithmicensure}{\textbf{Output:}}
	\caption{ESB-CPO}
	\label{alg:1}
	\begin{algorithmic}[1]
	    \STATE Orthogonal initialize the actor network and critic networks
	    \FOR {$k$ in 0, 1, 2, ...}
    	    \STATE Sample a set of trajectories $D = \{ \tau  \} \sim \pi_{\theta_k}$
            \FOR{$\tau$ in $D$}
                \FOR{$s$ in $\tau$}
                    \STATE Compute $\beta_{\theta_k}(s)$ with (\ref{beta})
                \ENDFOR
            \ENDFOR
            \STATE Compute $\alpha_{\theta_{k}}$ by solving local dual problem
            \STATE Form sample estimates $\hat{\mathfrak{g}}$, $\hat{\mathfrak{b}}$, $\hat{\mathcal{H}}$, $\hat{\mathfrak{c}}$ with $D$
            
    	    \IF {approximate ESB-CPO is feasible}
    	    \STATE Compute policy proposal $\hat{\theta}$ with (\ref{feasible})
            \ELSE
            \STATE Compute policy proposal $\hat{\theta}$ with (\ref{unfeasible})
    	    \ENDIF
        \STATE Obtain $\theta_{k+1}$ by backtracking line search to enforce satisfaction of constraint function in (\ref{ESB-CPO})
        \STATE Update critic networks by TD-like critic learning
	    \ENDFOR
	\end{algorithmic}  
\end{algorithm}

\begin{figure*}[ht!]
    \centering
    Average Returns:   
    
    \begin{subfigure}{\textwidth}
        \centering
        \vspace{0.22em}
        \begin{subfigure}{0.245\textwidth}
            \includegraphics[width=\textwidth]{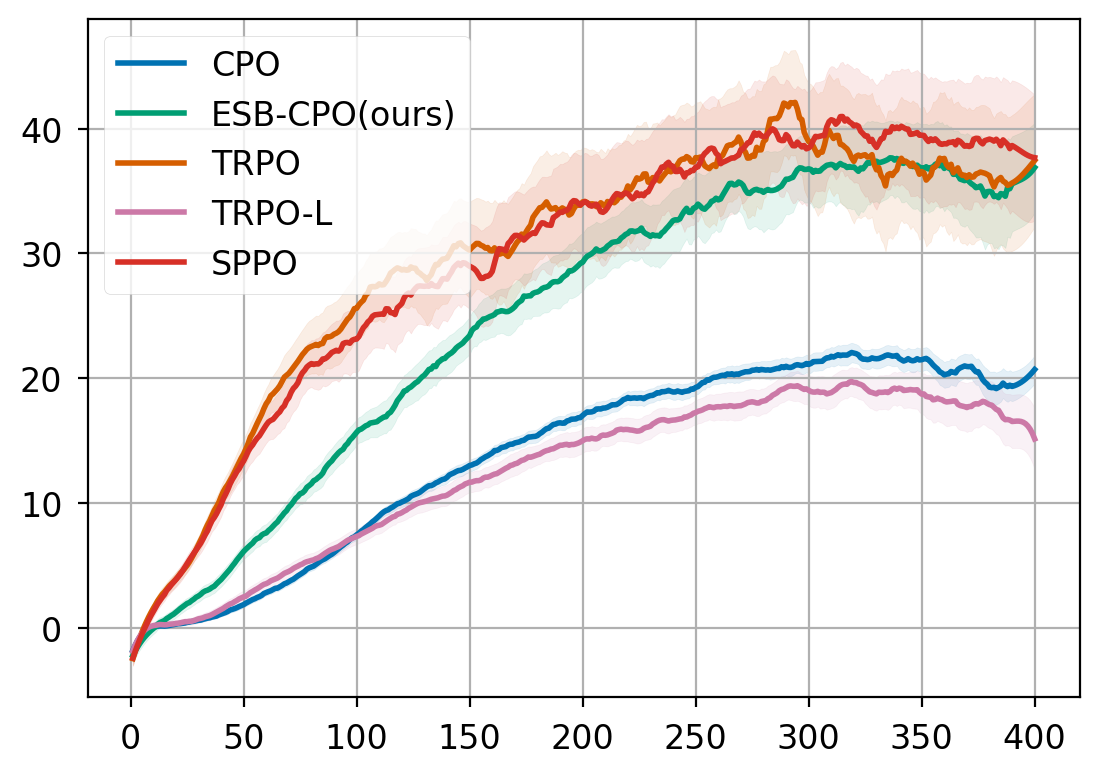}
        \end{subfigure}
        \begin{subfigure}{0.245\textwidth}
            \includegraphics[width=\textwidth]{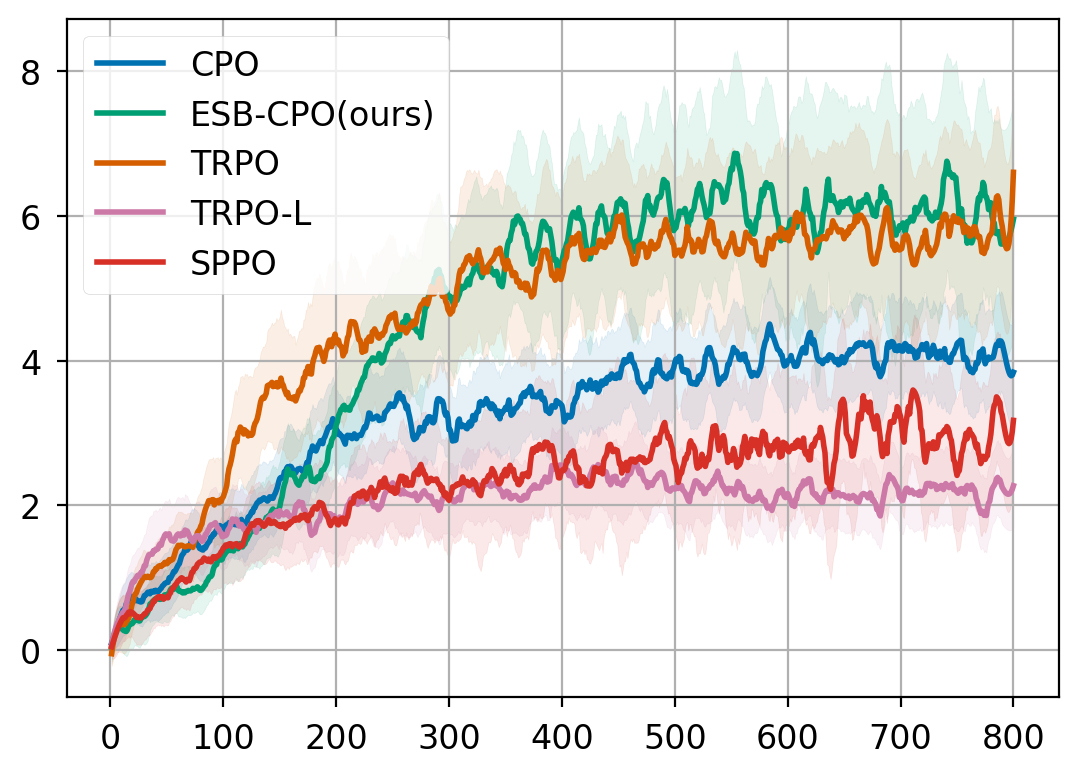}
        \end{subfigure}
        \centering
        \begin{subfigure}{0.245\textwidth}
            \includegraphics[width=\textwidth]{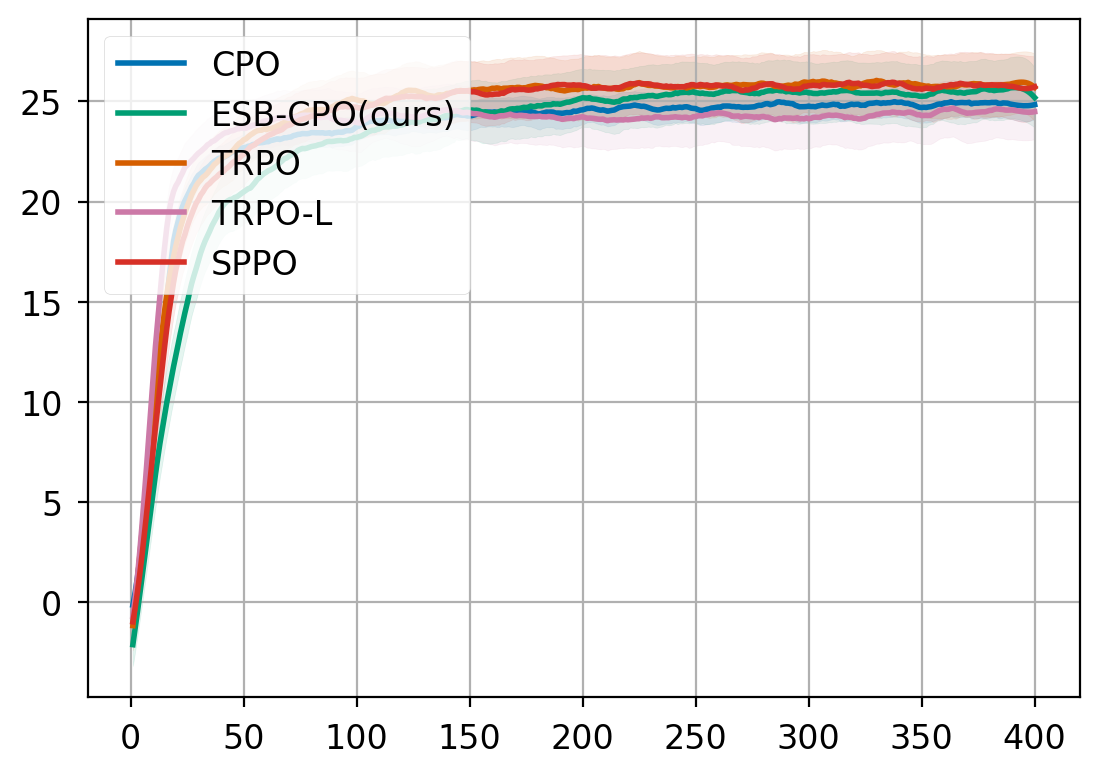}
        \end{subfigure}
        \begin{subfigure}{0.245\textwidth}
            \includegraphics[width=\textwidth]{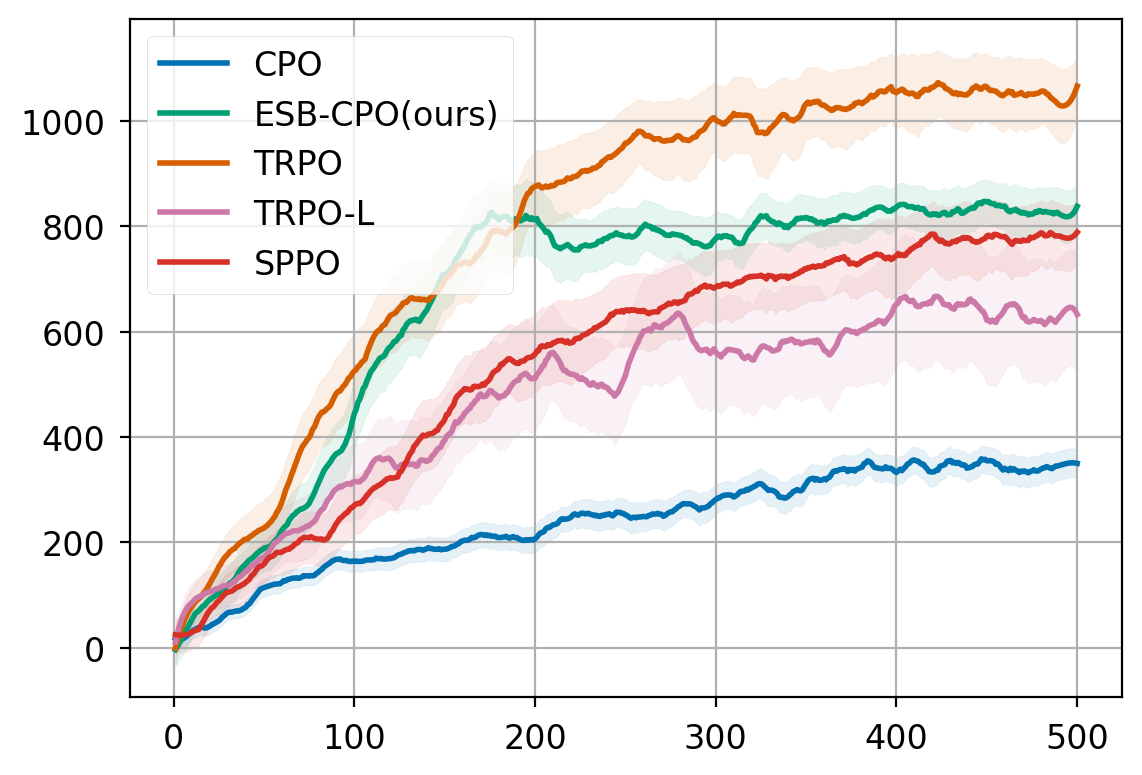}
        \end{subfigure}
    \end{subfigure}
    
    Average Costs:
    \begin{subfigure}{\textwidth}
        \centering
        \vspace{0.22em}
        \begin{subfigure}{0.245\textwidth}
            \includegraphics[width=\textwidth]{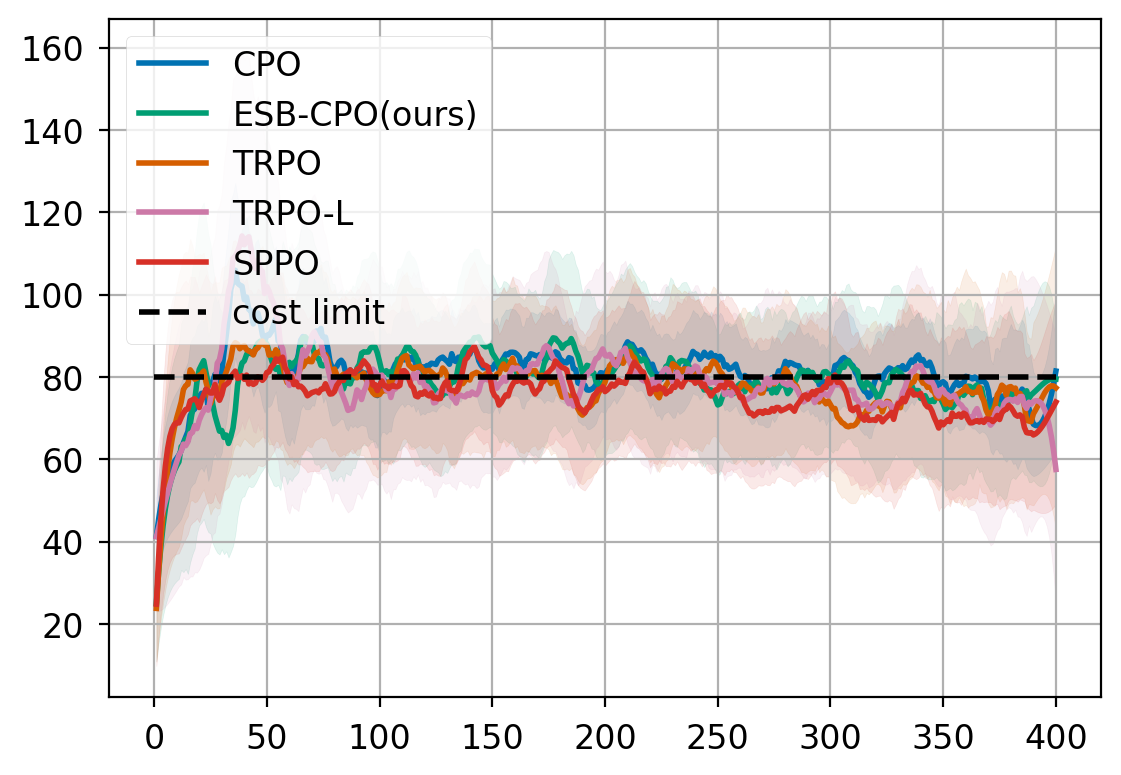}
            \caption{Doggo-Goal}
        \end{subfigure}
        \begin{subfigure}{0.245\textwidth}
            \includegraphics[width=\textwidth]{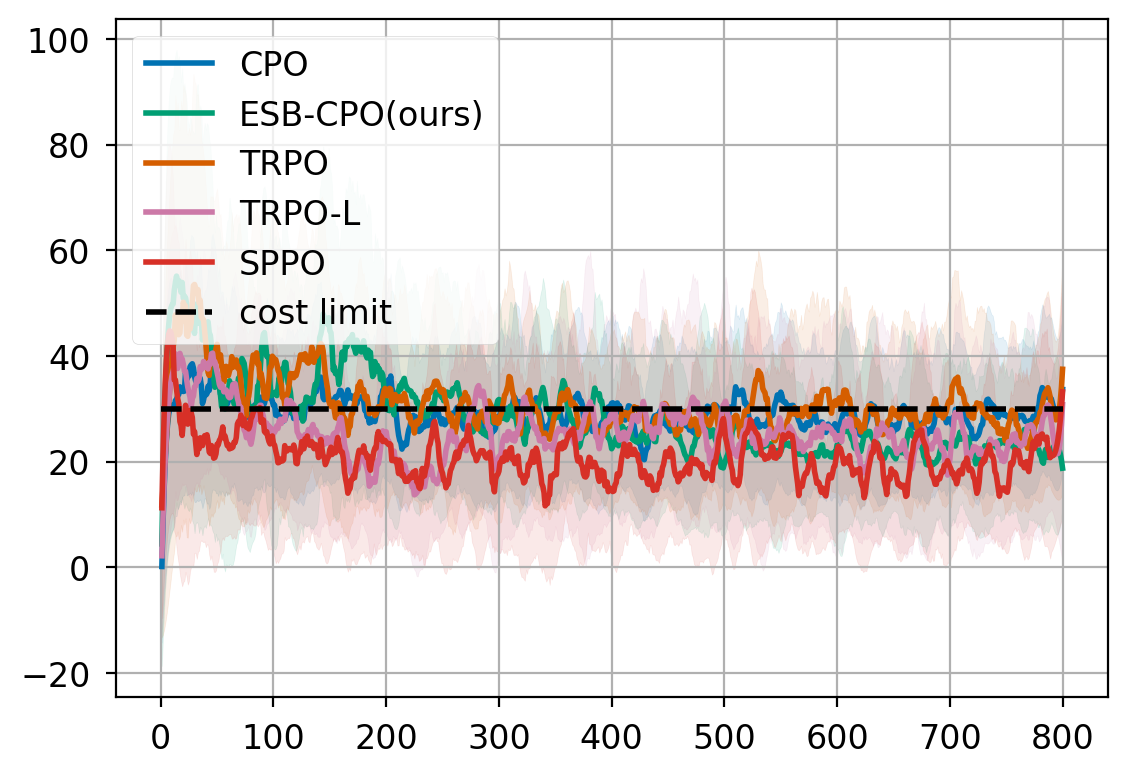}
            \caption{Car-Push}
        \end{subfigure}
        \begin{subfigure}{0.245\textwidth}
            \includegraphics[width=\textwidth]{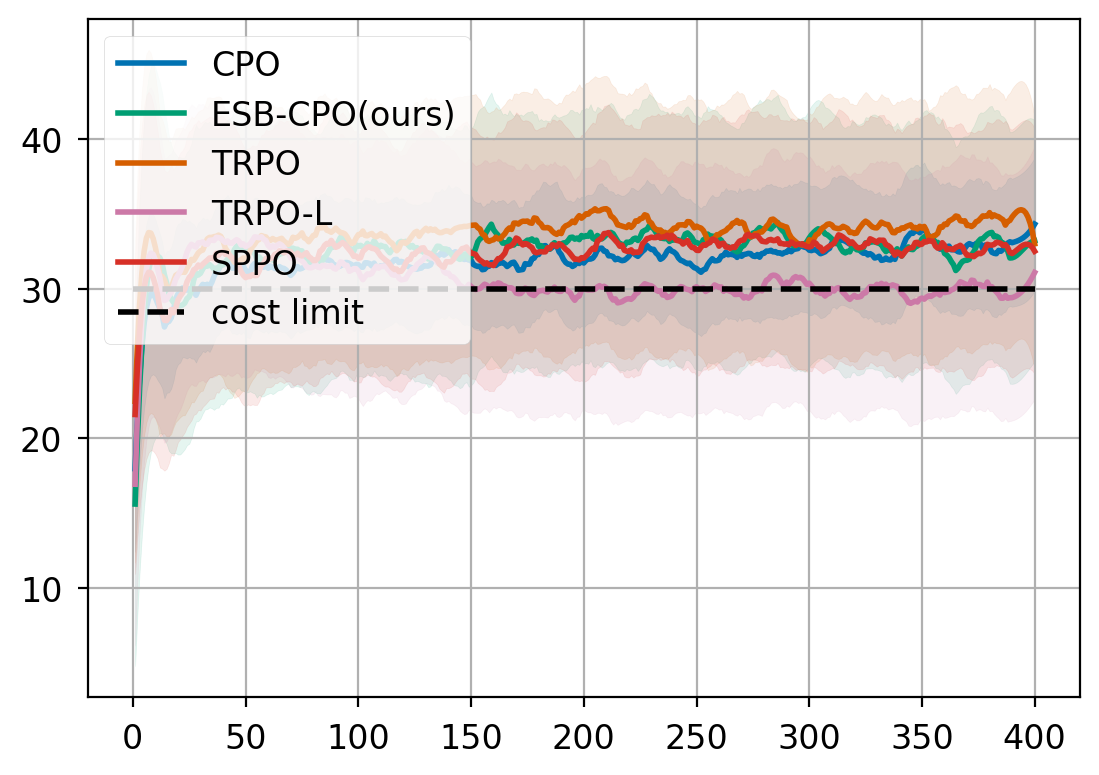}
            \caption{Ball-Reach}
        \end{subfigure}
        \begin{subfigure}{0.245\textwidth}
            \includegraphics[width=\textwidth]{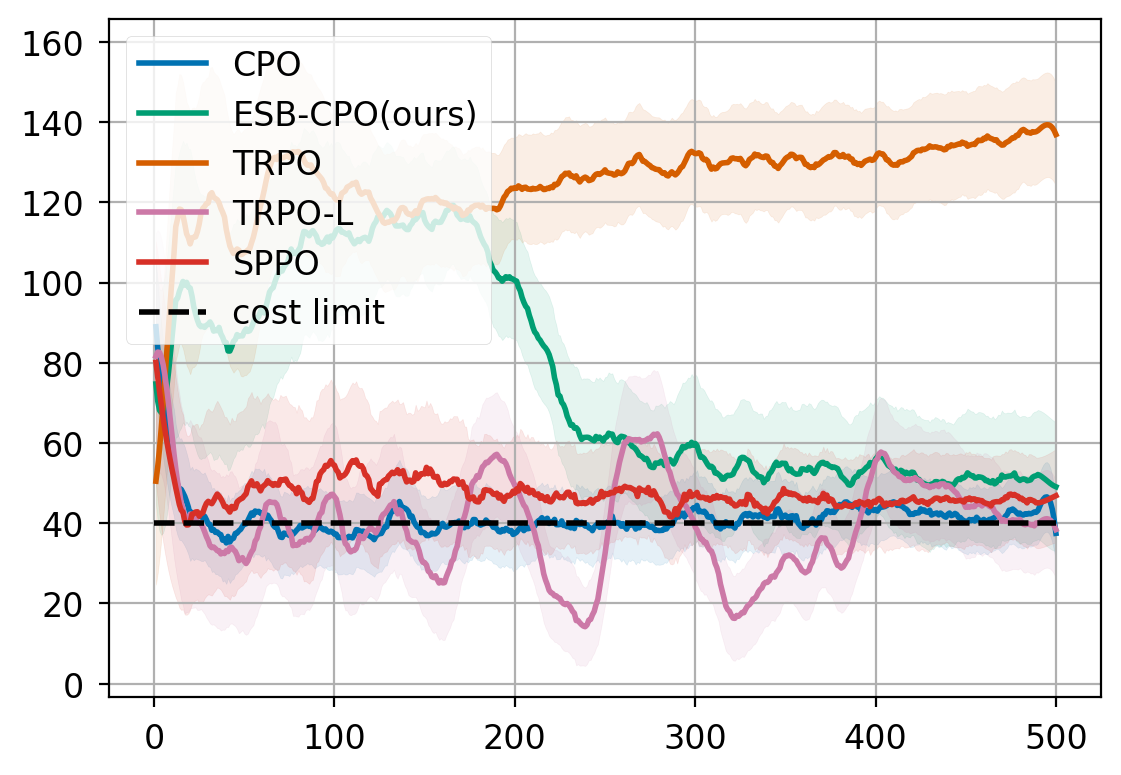}
            \caption{Drone-Circle}
        \end{subfigure}
    \end{subfigure}

    \caption{Average performance for ESP-CPO, CPO, SPPO, TRPO-L and TRPO over several seeds; the x-axis is training iteration. ESB-CPO outperforms the baselines in terms of average return or at least no worse than the best one of them, which proves that our method has better exploration efficiency. Though at the beginning of training ESB-CPO may fail to satisfy the constraints, it achieves satisfaction of constraints eventually or at least get very close to the limit during training.}
    \label{fig:results}
\end{figure*}

\begin{figure}[h]
    \centering
    \begin{subfigure}{0.235\textwidth}
        \centering
        \includegraphics[width=0.75\linewidth]{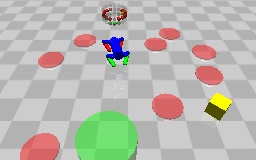}
        \caption{Doggo-Goal: a Safety-Gym task, where a quadruped robot need to navigation to a goal in an environment with barriers.}
    \end{subfigure}
    \begin{subfigure}{0.235\textwidth}
        \centering
        \includegraphics[width=0.75\linewidth]{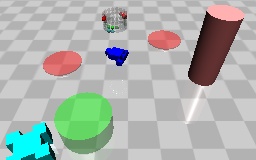}
        \caption{Car-Push: a Safety-Gym task, where a car-like robot need to push a box to a goal in an environment with barriers.}
    \end{subfigure}
    \begin{subfigure}{0.235\textwidth}
        \centering
        \includegraphics[width=0.75\linewidth]{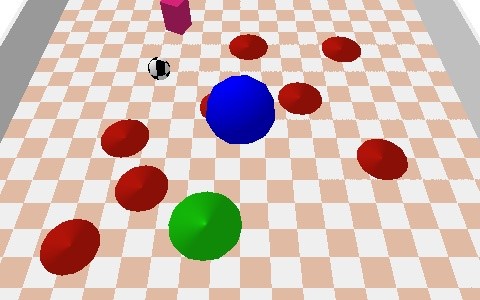}
        \caption{Ball-Reach: a Safety-Bullet-Gym task, where a spherical shaped robot need to reach a series of goals in an environment with barriers.}
    \end{subfigure}
    \begin{subfigure}{0.235\textwidth}
        \centering
        \includegraphics[width=0.75\linewidth]{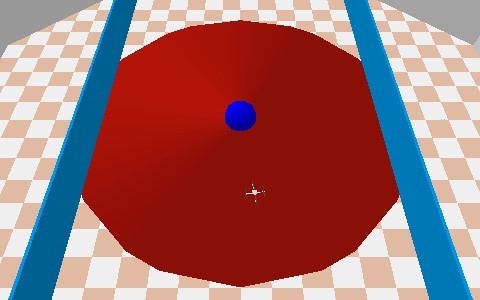}
        \caption{Drone-Circle: a Safety-Bullet-Gym task, where an air vehicle need to move on a circle in clock-wise direction and not go out of the safe region.}
    \end{subfigure}
    \caption{Specific tasks used in experimental part.}
    \label{fig:env}
\vspace{-10pt}    
\end{figure}

\section{EXPERIMENTS}
In this section, we design experiments to answer the following questions:
\begin{itemize}
    \item Does ESB-CPO outperform baseline algorithms on exploration efficiency?
    \item Does ESB-CPO achieve great satisfaction of constraint at the end of training?
    \item Does ESBs adaptively change as we expect?
\end{itemize}
\subsection{Comparison With Baselines}
We construct experiments on four tasks from two benchmarks, Bullet-Safety-Gym\cite{Gronauer2022BulletSafetyGym} and Safety-Gym\cite{Ray2019}. We give describes of tasks in Fig. \ref{fig:env}.
We use CPO\cite{achiam2017constrained}, SPPO\cite{chow2019lyapunovbased}, TRPO-Lagrangian\cite{peng2022model} as baselines \footnote{Baselines are implemented in \href{https://github.com/PKU-MARL/Safe-Policy-Optimization}{https://github.com/PKU-MARL/Safe-Policy-Optimization}}. These baselines are representative works of trust region based methods, Lyapunov-based methods and primal-dual methods, respectively. We also do experiments with TRPO\cite{schulman2015trust}, since our algorithm is developed from trust region theorems and TRPO is a good enough baseline to show the unconstrained situations. The results are shown in Fig. \ref{fig:results}.

In our experiments, ESB-CPO outperforms most of the baselines in total returns, and achieves good satisfaction of constraints. The training process of Drone-Circle significantly shows how our method works. In the early epochs, the returns and costs are both high and close to TRPO. The costs decrease to cost limit gradually. These results prove that in early epochs the agent explored efficiently with very loose constraints, and tried to avoid unsafe situations gradually. 
The results shows that our method allows overshoots of returns and violation of constraints in the early epochs, and constrains the policy to go back to a safe region eventually.

In the experiments of Doggo-Goal, we set a cost limit close to the average costs of TRPO, which means that it is almost constraint free. In this case we expect the policy to achieve a performance close to TRPO. The results show that ESP-CPO achieves the goal eventually, but some baselines have performance much worse than TRPO. These results prove that loosening constraints depending on constraints' satisfaction encourages exploration. 

\begin{figure}[h]
    \centering
    \begin{subfigure}{0.237\textwidth}
        \centering
        \includegraphics[width=\linewidth]{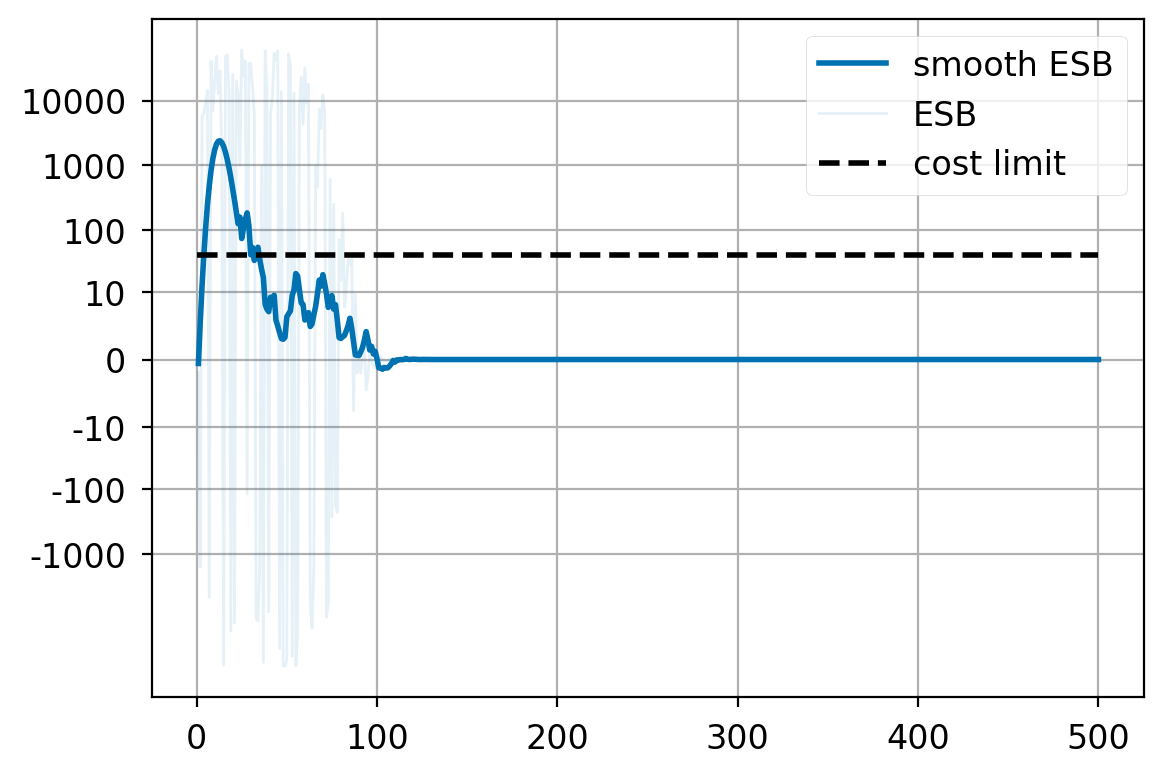}
        \caption{Drone-Circle}
    \end{subfigure}
    \begin{subfigure}{0.237\textwidth}
        \centering
        \includegraphics[width=\linewidth]{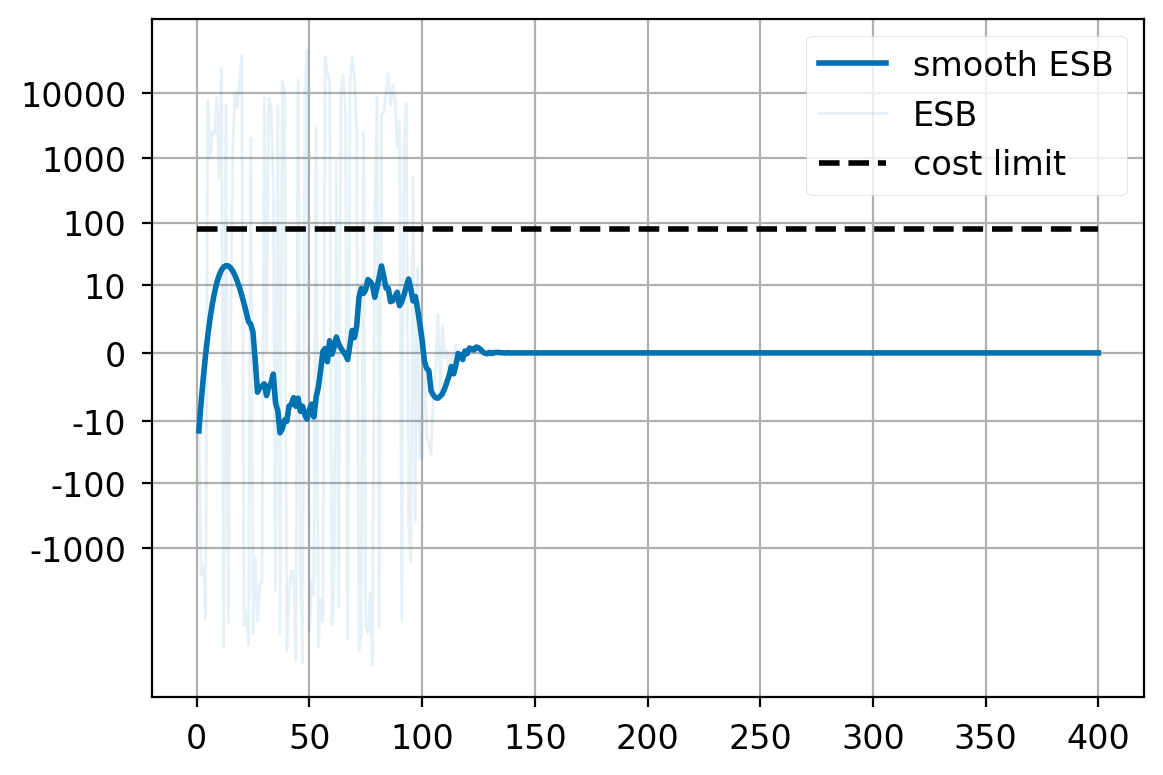}
        \caption{Doggo-Goal}
    \end{subfigure}
    \caption{Extra Safety Budgets in experiments.}
    \label{fig:exp-ESB}
\vspace{-10pt}   
\end{figure}

The total ESBs($-(G_{1\theta} (s, a) + G_{2\theta} (s, a))$) in the experiments of Drone-Circle and Doggo-Goal are shown in Fig. \ref{fig:exp-ESB}, which provide evidences that the constraints changed as what we expected. In the early epochs the ESBs greatly influence the total safety budget since their absolute value is much larger than cost limits; in the end, ESBs are close to 0 so that the policies are optimized to satisfy the original constraints. In most cases ESBs are positive. Thus ESBs loosen the constraints for better exploration efficiency. 

\begin{figure}[h]
    \centering
    Average Returns:
    \begin{subfigure}{0.5\textwidth}
        \centering
        \vspace{0.22em}
        \begin{subfigure}{0.45\textwidth}
            \includegraphics[width=\textwidth]{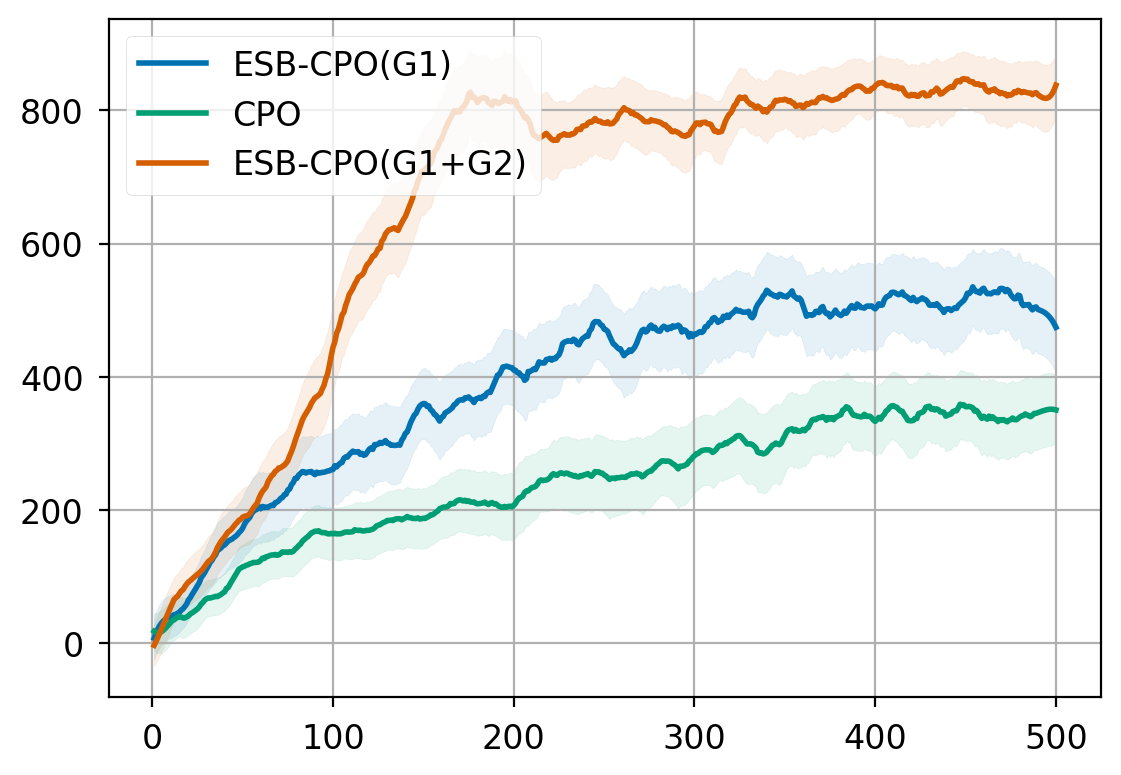}
        \end{subfigure}
        \begin{subfigure}{0.45\textwidth}
            \includegraphics[width=\textwidth]{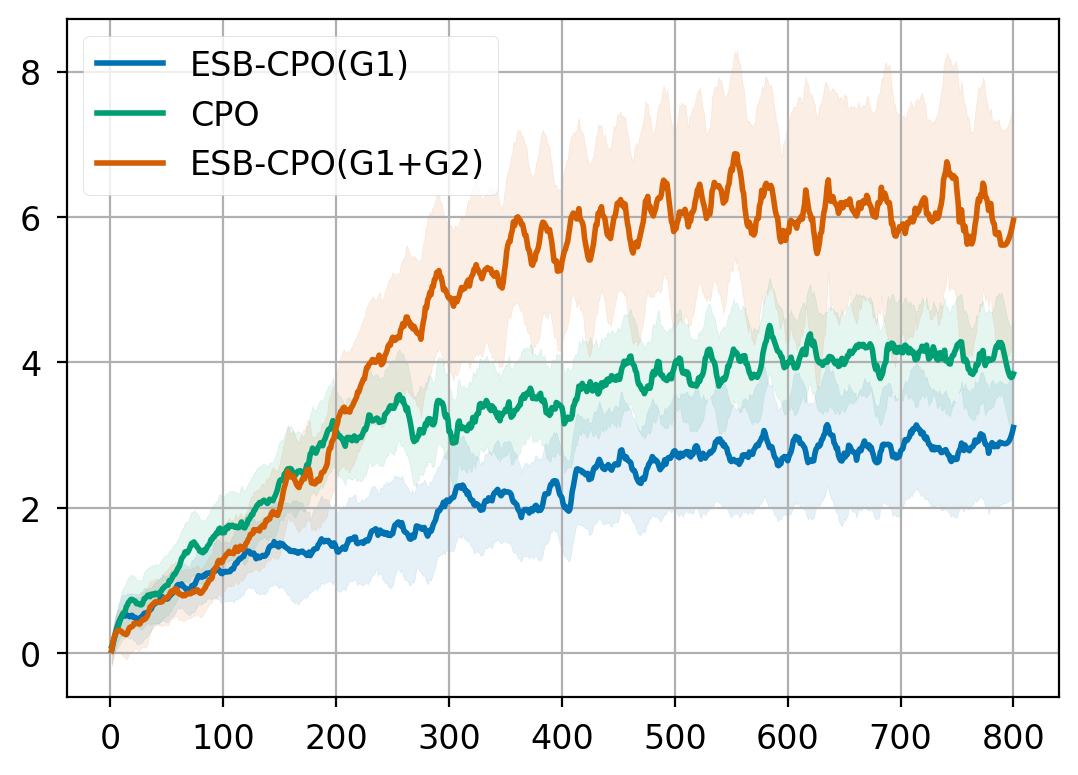}
        \end{subfigure}
    \end{subfigure}
    Average Costs:
    \begin{subfigure}{0.5\textwidth}
        \centering
        \vspace{0.22em}
        \begin{subfigure}{0.45\textwidth}
            \includegraphics[width=\textwidth]{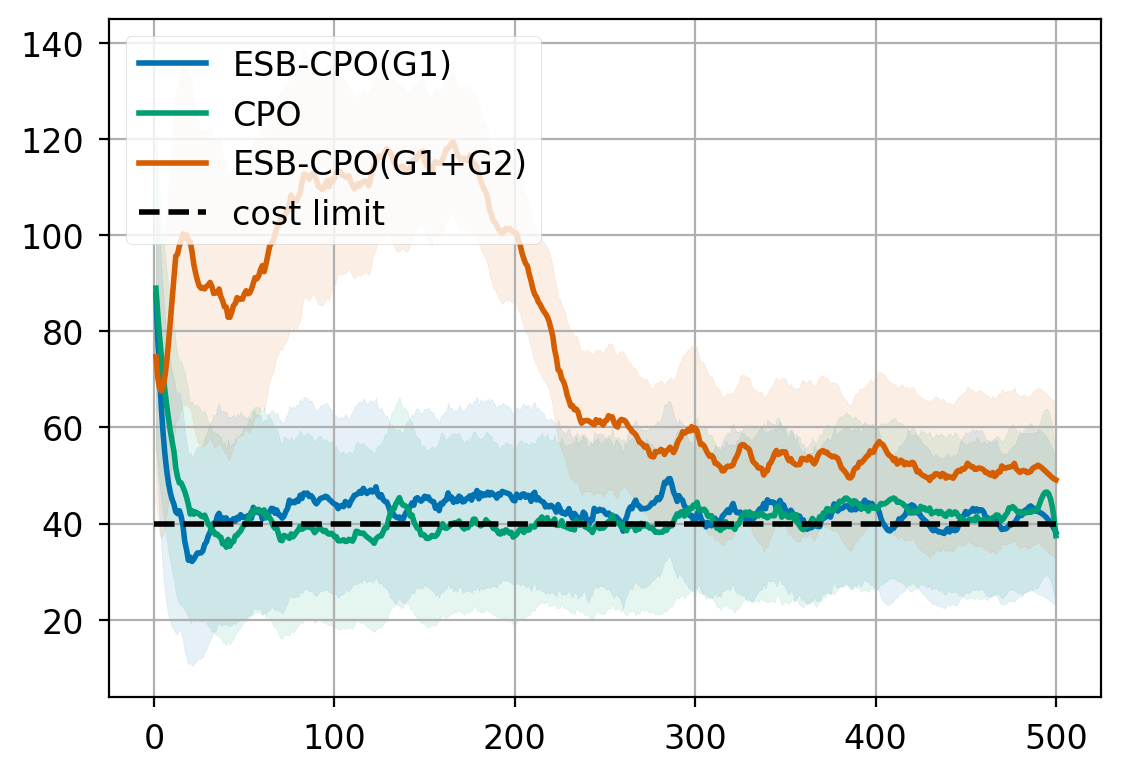}
            \caption{Drone-Circle}
        \end{subfigure}
        \begin{subfigure}{0.45\textwidth}
            \includegraphics[width=\textwidth]{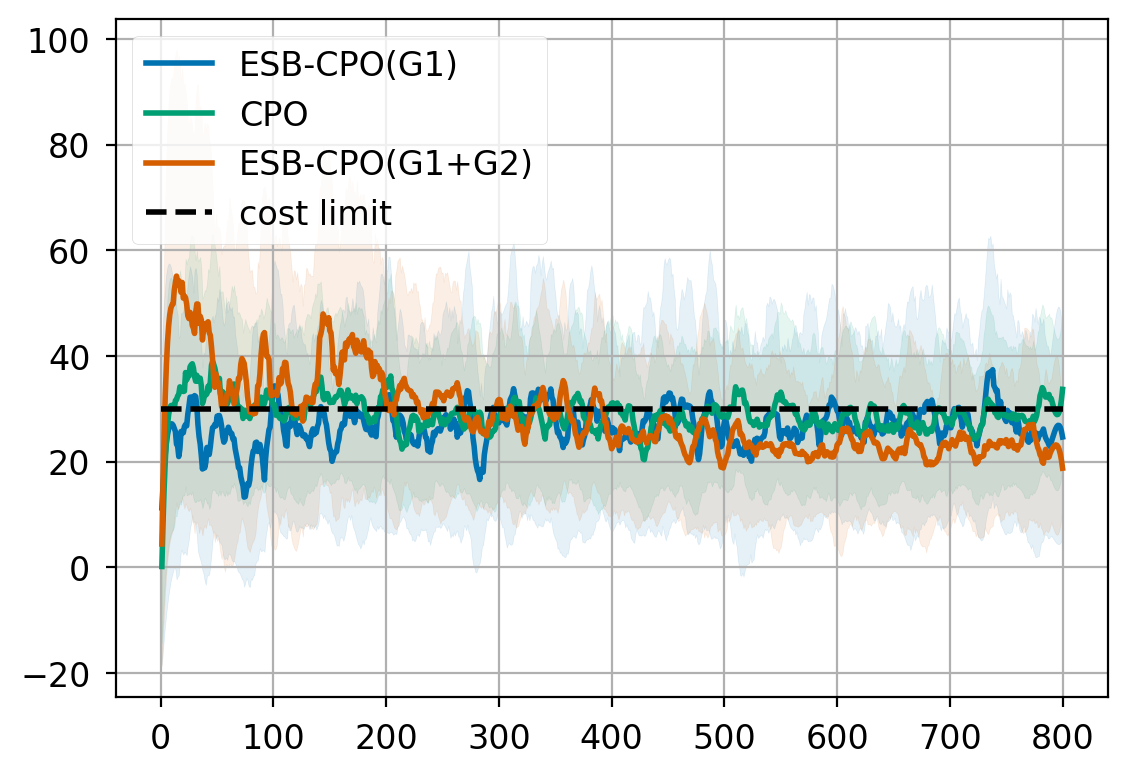}
            \caption{Car-Push}
        \end{subfigure}
    \end{subfigure}
    \caption{Average performance for ablation study. }
    \label{fig:ablation}
\vspace{-10pt} 
\end{figure}

\subsection{Ablation Study}
In Fig. \ref{fig:ablation}, we compare performance of CPO (ESB-CPO with no ESBs), ESB-CPO with only $G_{1\theta} (s, a)$ and full ESB-CPO (ESB-CPO with both $G_{1\theta} (s, a)$ and $G_{2\theta} (s, a)$). We denote the latter two algorithms as ESB-CPO (G1) and ESB-CPO (G1+G2), respectively.

$G_{1\theta} (s, a)$ controls the constraints based on stability, which is independent to training epochs. Thus the costs of ESB-CPO (G1) is similar to CPO, but have a slight gap since in some tasks stability is a tighter constraint. Stability is a task-dependent constraint, thus has different influences on the two tasks. Notice that in Car-Push, ESB-CPO (G1) has worse performance but slightly better satisfaction of constraint. 
 $G_{2\theta} (s, a)$ has a greater influence on constraint in early epochs to encourage exploration. Thus ESB-CPO (G1+G2) gains a significant improvement in return, though it has slightly higher costs than the other methods.

\section{CONCLUSIONS}

Constrained Policy Optimization with Extra Safety Budget (ESB-CPO) algorithm constructs a constrained optimizaiton problem based on trust region method. Different from CPO algorithm, we propose a new metric, namely Lyapunov-based Advantage Estimation (LAE) which consists of stability and safety values. It can magnify the gap between safe and unsafe transitions by safety value part.  When we view the safety value part as an extra safety budget, our method can loosen the constraints of unsafe transitions in the early stage. Meanwhile, our method can maintain the safety constraints gradually because the theoretical bound is very close to the bound in CPO algorithm. A promising direction of future work is to evaluate our method on more practical robotic tasks. Furthermore, we hope we can extend our work to off-policy and model-based RL methods.





\bibliographystyle{ieeetr} 
\bibliography{main}

\begin{thebibliography}{10}

\bibitem{hwangbo2019learning}
J.~Hwangbo, J.~Lee, A.~Dosovitskiy, D.~Bellicoso, V.~Tsounis, V.~Koltun, and
  M.~Hutter, ``Learning agile and dynamic motor skills for legged robots,''
  {\em Science Robotics}, vol.~4, no.~26, p.~eaau5872, 2019.

\bibitem{andrychowicz2020learning}
O.~M. Andrychowicz, B.~Baker, M.~Chociej, R.~Jozefowicz, B.~McGrew,
  J.~Pachocki, A.~Petron, M.~Plappert, G.~Powell, A.~Ray, {\em et~al.},
  ``Learning dexterous in-hand manipulation,'' {\em International Journal of
  Robotics Research (IJRR)}, vol.~39, no.~1, pp.~3--20, 2020.

\bibitem{garcia2015comprehensive}
J.~Garc{\i}a and F.~Fern{\'a}ndez, ``A comprehensive survey on safe
  reinforcement learning,'' {\em Journal of Machine Learning Research (JMLR)},
  vol.~16, no.~1, pp.~1437--1480, 2015.

\bibitem{yang2022safe}
T.-Y. Yang, T.~Zhang, L.~Luu, S.~Ha, J.~Tan, and W.~Yu, ``Safe reinforcement
  learning for legged locomotion,'' in {\em IEEE/RJS International Conference
  on Intelligent Robots and Systems (IROS)}, pp.~2454--2461, IEEE, 2022.

\bibitem{stooke2020responsive}
A.~Stooke, J.~Achiam, and P.~Abbeel, ``Responsive safety in reinforcement
  learning by pid lagrangian methods,'' in {\em International Conference on
  Machine Learning (ICML)}, pp.~9133--9143, PMLR, 2020.

\bibitem{liu2020ipo}
Y.~Liu, J.~Ding, and X.~Liu, ``Ipo: Interior-point policy optimization under
  constraints,'' in {\em Proceedings of the AAAI Conference on Artificial
  Intelligence (AAAI)}, vol.~34, pp.~4940--4947, 2020.

\bibitem{achiam2017constrained}
J.~Achiam, D.~Held, A.~Tamar, and P.~Abbeel, ``Constrained policy
  optimization,'' in {\em International Conference on Machine Learning (ICML)},
  pp.~22--31, PMLR, 2017.

\bibitem{Yang2020Projection-Based}
T.-Y. Yang, J.~Rosca, K.~Narasimhan, and P.~J. Ramadge, ``Projection-based
  constrained policy optimization,'' in {\em International Conference on
  Learning Representations (ICLR)}, 2020.

\bibitem{chow2018lyapunov}
Y.~Chow, O.~Nachum, E.~Duenez-Guzman, and M.~Ghavamzadeh, ``A lyapunov-based
  approach to safe reinforcement learning,'' {\em Advances in Neural
  Information Processing Systems (NeurIPS)}, vol.~31, 2018.

\bibitem{yang2023model}
Y.~Yang, Y.~Jiang, Y.~Liu, J.~Chen, and S.~E. Li, ``Model-free safe
  reinforcement learning through neural barrier certificate,'' {\em IEEE
  Robotics and Automation Letters (RAL)}, 2023.

\bibitem{liu2021policy}
Y.~Liu, A.~Halev, and X.~Liu, ``Policy learning with constraints in model-free
  reinforcement learning: A survey,'' in {\em International Joint Conference on
  Artificial Intelligence (IJCAI)}, 2021.

\bibitem{ding2021provably}
D.~Ding, X.~Wei, Z.~Yang, Z.~Wang, and M.~Jovanovic, ``Provably efficient safe
  exploration via primal-dual policy optimization,'' in {\em {International
  Conference on Artificial Intelligence and Statistics (AISTATS)}},
  pp.~3304--3312, PMLR, 2021.

\bibitem{Ray2019}
A.~Ray, J.~Achiam, and D.~Amodei, ``{Benchmarking Safe Exploration in Deep
  Reinforcement Learning},'' 2019.

\bibitem{peng2022model}
B.~Peng, J.~Duan, J.~Chen, S.~E. Li, G.~Xie, C.~Zhang, Y.~Guan, Y.~Mu, and
  E.~Sun, ``Model-based chance-constrained reinforcement learning via separated
  proportional-integral lagrangian,'' {\em IEEE Transactions on Neural Networks
  and Learning Systems}, 2022.

\bibitem{schulman2015trust}
J.~Schulman, S.~Levine, P.~Abbeel, M.~Jordan, and P.~Moritz, ``Trust region
  policy optimization,'' in {\em International Conference on Machine Learning
  (ICML)}, pp.~1889--1897, PMLR, 2015.

\bibitem{chow2019lyapunovbased}
Y.~Chow, O.~Nachum, A.~Faust, E.~Duenez-Guzman, and M.~Ghavamzadeh,
  ``Lyapunov-based safe policy optimization for continuous control,'' in {\em
  ICML 2019 Reinforcement Learning for Real Life Workshop}, 2019.

\bibitem{mathiesen2022safety}
F.~B. Mathiesen, S.~C. Calvert, and L.~Laurenti, ``Safety certification for
  stochastic systems via neural barrier functions,'' {\em IEEE Control Systems
  Letters (L-CSS)}, vol.~7, pp.~973--978, 2022.

\bibitem{yang2022cup}
L.~Yang, J.~Ji, J.~Dai, Y.~Zhang, P.~Li, and G.~Pan, ``Cup: A conservative
  update policy algorithm for safe reinforcement learning,'' {\em arXiv
  preprint arXiv:2202.07565}, 2022.

\bibitem{xu2020primal}
T.~Xu, Y.~Liang, and G.~Lan, ``A primal approach to constrained policy
  optimization: Global optimality and finite-time analysis,'' 2020.

\bibitem{srinivasan2020learning}
K.~Srinivasan, B.~Eysenbach, S.~Ha, J.~Tan, and C.~Finn, ``Learning to be safe:
  Deep rl with a safety critic,'' {\em arXiv preprint arXiv:2010.14603}, 2020.

\bibitem{sootla2022saute}
A.~Sootla, A.~I. Cowen-Rivers, T.~Jafferjee, Z.~Wang, D.~H. Mguni, J.~Wang, and
  H.~Ammar, ``Saut{\'e} rl: Almost surely safe reinforcement learning using
  state augmentation,'' in {\em International Conference on Machine Learning
  (ICML)}, pp.~20423--20443, PMLR, 2022.

\bibitem{sootla2022enhancing}
A.~Sootla, A.~I. Cowen-Rivers, J.~Wang, and H.~B. Ammar, ``Enhancing safe
  exploration using safety state augmentation,'' in {\em Advances in Neural
  Information Processing Systems (NeurIPS)} (A.~H. Oh, A.~Agarwal, D.~Belgrave,
  and K.~Cho, eds.), 2022.

\bibitem{chang2021stabilizing}
Y.-C. Chang and S.~Gao, ``Stabilizing neural control using self-learned almost
  lyapunov critics,'' in {\em IEEE International Conference on Robotics and
  Automation (ICRA)}, pp.~1803--1809, IEEE, 2021.

\bibitem{han2020actor}
M.~Han, L.~Zhang, J.~Wang, and W.~Pan, ``Actor-critic reinforcement learning
  for control with stability guarantee,'' {\em IEEE Robotics and Automation
  Letters (RAL)}, vol.~5, no.~4, pp.~6217--6224, 2020.

\bibitem{wang2023rl}
S.~Wang, F.~Lan, X.~Zheng, Y.~Cao, O.~Oseni, H.~Xu, Y.~Gao, and T.~Zhang, ``A
  rl-based policy optimization method guided by adaptive stability
  certification,'' {\em arXiv preprint arXiv:2301.00521}, 2023.

\bibitem{lawrence2020almost}
N.~Lawrence, P.~Loewen, M.~Forbes, J.~Backstrom, and B.~Gopaluni, ``Almost
  surely stable deep dynamics,'' {\em Advances in Neural Information Processing
  Systems (NeurIPS)}, vol.~33, pp.~18942--18953, 2020.

\bibitem{mittal2020neural}
M.~Mittal, M.~Gallieri, A.~Quaglino, S.~S.~M. Salehian, and J.~Koutn{\'\i}k,
  ``Neural lyapunov model predictive control,'' 2020.

\bibitem{dawson2022safe}
C.~Dawson, Z.~Qin, S.~Gao, and C.~Fan, ``Safe nonlinear control using robust
  neural lyapunov-barrier functions,'' in {\em Conference on Robot Learning
  (CoRL)}, pp.~1724--1735, PMLR, 2022.

\bibitem{murray2017mathematical}
R.~M. Murray, Z.~Li, and S.~S. Sastry, {\em A mathematical introduction to
  robotic manipulation}.
\newblock CRC press, 2017.

\bibitem{Gronauer2022BulletSafetyGym}
S.~Gronauer, ``Bullet-safety-gym: A framework for constrained reinforcement
  learning,'' tech. rep., mediaTUM, 2022.

\end{thebibliography}

\end{document}